\documentclass{article}

\usepackage{amsfonts}       
\usepackage{amsmath}
\usepackage{graphicx}

\usepackage{authblk}

\newcommand{\RR}{I\!\!R}
\newcommand{\EE}{\mathbb{E}}

\newcommand{\KL}{\mathrm{KL}}
\newcommand{\aopt}{\tilde{a}}

\newtheorem{theorem}{Theorem}
\newtheorem{lemma}[theorem]{Lemma}

\DeclareMathOperator*{\argmin}{arg\,min}

\title{The Combinatorial Multi-Bandit Problem \\and its Application to Energy Management\thanks{This project has received funding from the European Union’s Horizon 2020 research and innovation programme under grant agreement No 691735. The content of this article does not reflect the official opinion of the European Union. Responsibility for the information and views expressed in the article lies entirely with the authors.}}

\author[*]{Tobias Jacobs}
\author[*]{Mischa Schmidt}
\author[*]{S\'ebastien Nicolas}
\author[*]{Anett Sch\"ulke}
\affil[*]{ NEC Laboratories Europe, Heidelberg, Germany}
\date{}

\begin{document}

\maketitle

\begin{abstract}
  We study a \emph{Combinatorial Multi-Bandit Problem} motivated by applications in energy systems management. Given multiple probabilistic multi-arm bandits with unknown outcome distributions, the task is to optimize the value of a combinatorial objective function mapping the vector of individual bandit outcomes to a single scalar reward.   
  Unlike in single-bandit problems with multi-dimensional action space, the outcomes of the individual bandits are observable in our setting and the objective function is known. 
Guided by the hypothesis that individual observability enables better trade-offs between exploration and exploitation, we generalize the lower regret bound for single bandits, showing that indeed for multiple bandits it admits parallelized exploration.
For our energy  management application we propose a range of algorithms that combine exploration principles for multi-arm bandits with mathematical programming. In an experimental study we demonstrate the effectiveness of our approach to learn action assignments for 150 bandits, each having 24 actions, within a  horizon of 365 episodes.
\end{abstract}

\section{Introduction}

We study a cooperative multi-agent learning problem motivated by applications in energy systems management. Consider an interconnected system with energy producers, consumers, and storage components. Management units are responsible to steer the behavior of different subsets of system components, contributing to the overall goal to balance energy supply and demand while keeping the system state within its operational constraints. Large-scale energy systems management is typically realized in a hierarchical fashion with higher-level management distributing and delegating targets.

From the viewpoint of a particular management unit, a multitude of~$n$ system components, or subsystems, is to be operated. Each component can receive a range of instructions (e.g. management policies), and the outcome of the instruction can be observed (e.g. by the load curve of the component). Among the possible management goals are balancing supply and demand, reducing peak load, and reducing consumption during a given target time interval. While the level of fulfillment of the management goal can be calculated from the observed behavior of the involved system components, the influence of the instructions on the behavior follows a probability distribution that is  unknown to the management unit and can only be learned by experience.

The above management task is an application of a general multi-bandit problem. Each system component corresponds to a multi-arm bandit or \emph{actor}, the range of instructions corresponds to the arms or \emph{actions} of the bandit, and the observed behavior is the stochastic result of pulling an arm. 
The reward is the level of fulfillment of the management goal, which can be computed by a  known function mapping the~$n$ behavior observations (e.g. load curves) to a scalar. 


\paragraph{Our contribution}

To the best of our knowledge we are the first to study the variant of the combinatorial bandit problem where the reward function combines the output of several bandits. The formal problem definition is given in Section~\ref{sec:definition}. We generalize the lower regret bound of Lai and Robbins~\cite{lai85} to the Combinatorial Multi-Bandit scenario. The generalized bound is established in Section~\ref{sec:lowerbound} by the requirement to apply suboptimal actions of all actors with a distribution-dependent frequency, but with the possibility to do this in parallel for different actors. We describe how the problem is applied to energy systems management in Section~\ref{sec:management} and give a range of action selection algorithms combining bandit exploration schemes with mathematical programming. In an experimental study presented in Section~\ref{sec:experiments}, the  algorithms are applied to optimize a load shifting objective in a setup with 150 energy consumers. The experiments demonstrate the ability of our approach to parallelize the exploration process, finding good action assignments within 365 episodes.

The Combinatorial Multi-Bandit Problem is applicable beyond energy management. Coordinating a multitude of unknown and heterogeneous actors is an essential requirement for effective systems management, including but not limited to logistics, robotics, communication networks, and cloud computing systems. Although such tasks can in principle also be addressed by traditional Reinforcement Learning methods with multi-dimensional action space and a scalar reward signal, doing so comes at the cost of discarding the individual feedback of each managed component. Combinatorial bandits offer a more fine-grained way to model these management problems, thus potentially speeding up the learning and optimization process. Although the scope of this work is restricted to stateless systems, the problem can also be generalized to contextual bandits or even full MDPs.

\paragraph{Related work}

The authors of~\cite{gabillon11} have studied a multi-bandit problem where in each episode only one bandit can apply an action. The challenge is to distribute a limited budget of exploration episodes among the bandits to determine the best action for each of them with sufficient certainty.

The variant most similar our scenario is the \emph{Combinatorial Multi-Armed Bandit} problem introduced by Chen et al. in~\cite{chen13}. Here multiple arms of a single bandit can be played in each episode, where the set of admissible arm combinations is limited by arbitrary constraints. Each arm, whether played or not, produces some outcome in each episode, and the reward is only depending on these outcomes. Among the differences to our Combinatorial Multi-Bandit Problem are that (a) we consider multiple bandits each of which has to select exactly one of its arms, (b) unlike~\cite{chen13} our analyses do not assume any properties (like monotonicity, smoothness) of the reward function.

For the variant of the Combinatorial Multi-Armed Bandit Problem with full observability of  individual arm outcomes, Durand and Gagn\'e have generalized the Thompson sampling technique and applied it to Online feature selection in~\cite{durand14}. In~\cite{combes15} Combes et al. study a version of the problem with linear reward functions that was introduced as \emph{Combinatorial Bandits} in~\cite{cesa12}, providing tight lower bounds for the regret and providing action selection rules using the upper confidence bound method for exploration. Algorithms for the case of linear reward functions have also been proposed in~\cite{gai12}.

\section{Problem definition and notation}
\label{sec:definition}


In the \emph{Combinatorial Multi-Bandit Problem} we are given a multitude of~$n$ actors, indexed by~$i= 1 \ldots n$. 
Each actor~$i$ can perform a number~$k^i$ of different actions, indexed by~$a^i =1 \ldots k^i$. Selecting an action~$a^i$ for actor~$i$ results in a probabilistic output~$o^i \in O^i$. Unlike in most of the related work, the outputs~$o^i$ do not directly represent scalar rewards but can be elements of any space~$O^i$ over which probability measures can be defined. For each actor~$i$ the distribution~$\mathbb{P}(o^i)$ over the possible outputs is unknown and only depends on the most recent action~$a^i$ that has been chosen for the actor. What is known is the \emph{objective function}~$r:O^1 \times \ldots \times O^n \to \RR$ which combines the individual actor outputs into a scalar \emph{reward}.

Time proceeds in discrete episodes~$t=1\ldots T$. In each episode an \emph{action assignment}~$a_t = (a^1_t, \ldots, a^n_t)$ has to be made, assigning to each actor~$i$ an action~$a^i_t$. We also use vector notation~$o = (o^1,\ldots,o^n)$ to denote the  outputs of the~$n$ actors observed at the end of each episode.
The objective is to maximize the \emph{total expected reward}
$
	\sum_{t=1}^T \mathbb{E}[r(o) \mid a_t] \ .
$

As the outcomes only depend on the actions chosen in the same episode, there exists an optimal action assignment~$\aopt = (\aopt^{1},\ldots,\aopt^{n})$ that maximizes the expected reward independently of the current episode~$t$. If~$\aopt$ was known, then a trivial optimal policy would be to select~$\aopt$ in each episode. 

As $\aopt$ and the outcome probabilities are unknown, an optimal or near-optimal assignment of actions has to be learned from experience using a strategy that balances \emph{exploration} (choosing actions for the purpose of learning their outcome distribution) and \emph{exploitation} (choosing action assignments known to perform well). When applying a particular action assignment~$a = (a^1,\ldots,a^n)$, the expected gap compared to the optimal assignment~$\aopt$ is called the \emph{regret} 
\begin{equation}
\label{eq:regret}
\rho(a) = \mathbb{E}[r(o) \mid \aopt] - \mathbb{E}[r(o) \mid a] \ ,
\end{equation}
and the \emph{normalized regret} is defined as~$\rho(a) / \mathbb{E}[r(o) \mid \aopt]$.
Given action assignments~$a_1,\ldots,a_t$, the \emph{cumulative regret} until episode~$t$ is defined as 
\begin{equation}
R(t) = \sum_{t^\prime = 1}^{t} \rho(a_{t^\prime}) = t \cdot \mathbb{E}[r(o) \mid \aopt] - \sum_{t^\prime = 1}^{t} \mathbb{E}[r(o) \mid a_{t^\prime}] \ ,
\end{equation}
and the \emph{cumulative normalized regret} is defined accordingly. 

\section{Lower regret bound}
\label{sec:lowerbound}

In~\cite{lai85} it was proven that for multi-arm bandit problems with action space~$A$, optimal action~$\aopt$ and reward~$r$, the cumulative regret until episode~$t$ of any action selection rule satisfies
\begin{equation}
\label{eq:lairobbins}
	\lim_{t \to \infty} \frac{R(t)}{\log t} \geq \sum_{a \in A} \frac{\EE[r \mid \aopt] - \EE[r \mid a]]}{\KL(a,\aopt)} \ ,
\end{equation}
under fairly non-restrictive conditions of the family of possible reward distributions. In the formula above,~$\KL(a,\aopt)$ is the Kullback-Leibler divergence of the reward distributions of~$a$ and~$\aopt$. 


This lower bound applies when interpreting the Combinatorial Multi-Bandit Problem with~$n$ actors as a standard multi-arm bandit problem with a single actor having an~$n$-dimensional action space.
Each action of the single actor corresponds to an action assignment to the~$n$ original actors, and the result of the reward function is used as the reward signal. Using this interpretation, known algorithms for multi-arm bandit problems can be directly applied, but the  number of actions is exponential in $n$, and 
Equation~\ref{eq:lairobbins} applies with exponentially many summands.

Our generalization of Equation~\ref{eq:lairobbins} is established following a similar argumentation as the proof given in~\cite{lai85}. In that work it has been shown that any action selection strategy either satisfies the bound in a trivial way or applies, with probability~$1$, each suboptimal action~$a$ at least~$\KL(a,\aopt)^{-1} \log t$ times as~$t$ goes to infinity. An analog statement can be made for our multi-bandit problem, introducing~$\eta(a,t)$ to define the number of times the action selection rule under consideration has applied action assignment~$a$ in the first~$t$ episodes. Accordingly,~$\eta^i(a^i,t)$ is the number of times actor~$i$ has applied action~$a^i$. We say that action $a^i$ for actor $i$ is suboptimal when it is not part of any optimal action assignment. As the proof of the following lemma follows the same argumentation as the proof of Theorem 2 in~\cite{lai85}, we only provide a proof sketch here.

\begin{lemma}
\label{lemma:1}
Assume that an action selection rules satisfies
\begin{equation}
\label{eq:precondition}
\mathbb{E}[\eta(a,t)] = o(t^\alpha)
\end{equation}
for any~$\alpha>0$ and each suboptimal action assignment~$a$.  Then, for any actor~$i$ and any suboptimal action~$a^i$, it holds that
\begin{equation}
\label{eq:consequence}
\lim_{t \to \infty} \mathbb{P} \Bigl(\eta^i(a^i,t) <\frac{\log t}{\KL(a^i,\aopt^i)}\Bigr) = 0 \ ,
\end{equation}
where~$\aopt^i$ is the action of actor~$i$ in an optimal assignment.
\end{lemma}

\paragraph{Proof sketch}
 Fix some suboptimal action~$a^i$ and let~$\aopt^i$ be the action assigned to actor~$i$ by an optimal assignment. In the problem instance under consideration, let~$\Phi$ be the outcome probability distribution of~$a^i$. We define a modified instance, where the outcome distribution of~$a^i$ is set to a distribution~$\Phi^\prime$ infinitesimally close to the outcome distribution of~$\aopt^i$, but defined in a way that~$a^i$ instead of~$\aopt^i$ becomes part of the optimal assignment under~$\Phi^\prime$.

For convenience of notation, define condition~$C_t$ as~$\eta^i(a^i,t) <\log t \cdot \KL(a^i,\aopt^i)^{-1}$.
Applying the precondition from Equation~\ref{eq:precondition} to the modified instance with~$\Phi^\prime$, we obtain that~$\mathbb{E}_{\Phi^\prime}[\eta^i(a^i,t)] = t - o(t^\alpha)$ for any~$\alpha>0$. From this lower bound on the expectation of~$\eta^i(a^i,t)$ under~$\Phi^\prime$ we can apply the Markov inequality to obtain an upper bound of roughly~$1/t$ on~$\mathbb{P}_{\Phi^\prime}(C_t)$. 

For Equation~\ref{eq:consequence} to hold for the original problem instance, it suffices to show that 
\begin{equation}
\label{eq:relation}
\mathbb{P}_{\Phi}(C_t) \to t^{1-\epsilon} \mathbb{P}_{\Phi^\prime}(C_t) \ \mathrm{for} \ t \to \infty
\end{equation}
for some~$\epsilon>0$. To establish this, note that~$\mathbb{P}_{\Phi}(C_t)$ and~$\mathbb{P}_{\Phi^\prime}(C_t)$ are calculated by integrating over the probability densities of all sequences of $t$ episodes with $C_t$ satisfied. In any such sequence~$Y$, let $O$ be the multi-set of outputs observed by actor $i$ after action $a^i$ has been applied. Due to $C_t$, $|O| < \log t \cdot \KL(a^i,\aopt^i)^{-1}$. As the only difference between $\mathbb{P}_{\Phi}(Y)$ and $\mathbb{P}_{\Phi^\prime}(Y)$ is in terms of the output probabilities of the elements of $O$, it follows that $\mathbb{P}_{\Phi}(Y) = \mathbb{P}_{\Phi^\prime}(Y) \cdot \Pi_{o \in O} \mathbb{P}_{\Phi}(o) / \mathbb{P}_{\Phi^\prime}(o)$. The divergence $\KL(a^i,\aopt^i)$ is the expected logarithm of $\mathbb{P}_{\Phi^\prime}(o) / \mathbb{P}_{\Phi}(o)$; thus it can be shown that the factor $\Pi_{o \in O} \mathbb{P}_{\Phi}(o) / \mathbb{P}_{\Phi^\prime}(o)$ almost surely converges to $t^{1 - \epsilon}$.
%
%
%
%
%
%
\hfill{$\Box{}$}

Lemma~\ref{lemma:1} establishes that, in order to attain sub-polynomial regret, the assignments have to be such that each suboptimal action~$a^i$ is applied by its actor~$i$ with a frequency that converges to~$\log t \cdot \KL(a^i,\aopt^i)^{-1}$. The action~$a^i$ can appear in a range of different suboptimal action assignments, possibly causing different regrets. The lower regret bound represents the best selection of assignments that apply all suboptimal actions sufficiently often.

Formally, for~$i=1 \ldots n$, let~$A^i$ be the action set of actor~$i$, and let~$A = A^i \times \ldots \times A^n$ be the set of all assignments. We define a \emph{mixed assignment} as a weight function~$w:A \to \RR^+_0$. The weighting is called \emph{feasible} if for each action~$a^i \in A^i$ of any actor~$i$ the total weight of all assignments assigning~$a^i$ to actor~$i$ is at least~$\KL(a^i,\aopt^i)^{-1}$, that is, if~$\sum_{a^i \in a} w(a) \geq \KL(a^i,\aopt^i)^{-1}$. The regret~$\rho(a)$ of an assignment~$a \in A$ is defined as in Equation~\ref{eq:regret}, and the regret of a mixed assignment~$w$ is defined~as 
$
\rho(w) := \sum_{a \in A} w(a) \rho(a) 
$.
We define
\begin{equation}
w^* := \argmin_ {w \ \mathrm{feasible}} \rho(w) \ .
\end{equation}

\begin{theorem}
\label{thm:lowerbound}
For any action selection rule for the Combinatorial Multi-Bandit Problem the cumulative regret almost surely satisfies
$$
\lim_{t \to \infty} \frac{R(t)}{\log t} \geq \rho(w^*) \ .
$$
\end{theorem}

This lower bound is a generalization of the lower bound for single bandits as stated in Equation~\ref{eq:lairobbins}. To see this, observe that, when~$n=1$, it holds that~$A = A^1$ and the feasibility condition on~$w^*$ implies that~$w^*(a) = \KL(a,\aopt)^{-1}$ for any~$a \in A$.

\section{Multi-bandit algorithms for energy systems management}
\label{sec:management}

Our study of the Combinatorial Multi-Bandit Problem is motivated by an application in the domain of energy management in a Smart City project. 
In this project, an \emph{Energy Demand Management System} is coordinating consumers of energy with the goal to reduce consumption when required by the network.
This and other use cases of \emph{demand-side management} are gaining importance in the energy sector due to the increasing share of volatile renewable energy sources~\cite{haider16}.

When modeling this application as a multi-bandit problem, each consumer corresponds to an actor, and the action space consists of signals that can be sent to energy consumers on a daily basis. The signals incentivize consumers to re-schedule or curtail their energy consumption in various ways. At the end of each day the load curves of the consumers are observed. A metric, defined on the aggregated load curve of all consumers, measures the extent to  which the energy management goal has been fulfilled. In this setting we consider the target to obtain the maximal load reduction that can be sustained during a given \emph{target time interval}, measured against a given baseline curve. This specification models a common use case in energy demand management, where e.g. the owner of a management unit agrees with the network operator to provide load reduction as a service; see~\cite{siano14}.

Formally, there are~$n$ actors, here each having the same number~$k$ of actions. Having applied an action, the outcome is observed as an~$H$-dimensional discrete \emph{load reduction curve}~$l^i = (l^i_1,\ldots,l^i_H)$. The objective function~$r$ is defined as 
\begin{equation}
\label{eq:objective}
	r(l^1,\ldots,l^n) = \min_{h = 1 \ldots H} \sum_{i = 1}^n l^i_h \ .
\end{equation}

\paragraph{Algorithmic framework}

For the application just described we design a range of algorithms that combine combinatorial optimization with exploration strategies for bandit problems. In the next paragraphs we describe the optimization module used by all algorithms, and subsequently we explain how the different algorithms apply the module.

\paragraph{Optimization module}

Assume that complete information about the environments for $N \geq 1$ \emph{sample episodes}  is available, that is, for each~$t = 1 \ldots N$ a known function 
$
	\ell_t(i,j,h)
$
specifies the load reduction that will be achieved at any daytime~$h$ when applying any action~$j$ to any actor~$i$. Given this information, the optimization module computes an action assignment maximizing the \emph{average} reward over the~$N$ sample episodes.

This computation is realized by formulating and solving the problem to maximize the average result of Equation~\ref{eq:objective} as an integer linear problem. The program uses~$n \cdot k$ binary variables~$b_{ij}$ to specify whether actor~$i$ will apply action~$j$. 
\begin{equation}
\begin{split}
\mathrm{maximize} \ \
&
 \sum_{t = 1}^N M_t
\\
\mathrm{subject\ to} \ \
&
\sum_{i = 1}^n \sum_{j=1}^k \ell_t(i,j,h) \cdot b_{ij} \geq M_t \ \ \ \ \mathrm{for} \ h = 1 \ldots  H, t = 1,\ldots,N
\\
{}
&
\sum_{j = 1}^k b_{ij} = 1
\hspace{2.53cm}\ \ \ \mathrm{for} \ i=1 \ldots n
\\
{}
& 
b_{ij} \in \{0,1\}
\hspace{2.53cm} \ \ \  \mathrm{for} \ i = 1 \ldots n, j = 1 \ldots k \ .
\end{split}
\end{equation}
Each summand~$M_t$ of the objective function represents the reward obtained in one episode~$t$. For any episode~$t$ and any daytime~$h$ within that episode, an upper bound on~$M_t$ is given by the total load reduction of all actors at that daytime. Thus, for each episode~$t$ there are~$H$ such upper bounds on~$M_t$. These~$N \cdot H$ constraints are represented by the second line of the program. The third line represents the~$n$ constraints that only one action can be chosen for each actor~$i=1 \ldots n$.

We remark that the mathematical program cannot be solved in polynomial worst case runtime unless~$\mathrm{P}=\mathrm{NP}$. To see this, consider the special case with~$N=1$,~$k=2$ and~$\ell_1(i,j,h) \in \{0,1\}$ for all~$i=1 \ldots n$,~$j \in {1,2}$,~$h = 1 \ldots H$. The problem to decide whether there is an action assignment with reward at least~$1$ is equivalent to the SAT problem with~$n$ variables and~$H$ clauses, which is known to be NP-complete~\cite{karp72}.

Nevertheless, state-of-the-art solvers for integer programs can solve fairly large instances within reasonable runtime. When computation time runs out, solvers can also return the best suboptimal solution found so far, including an upper bound on the remaining optimality gap.

\paragraph{Single Episode algorithm}

We present a \emph{Single Episode (SE)} algorithm using the optimization module with~$N=1$ to compute an optimal action assignment for a single episode. This episode, specified by function~$\ell := \ell_1$, has to represent the population of all possible episodes. 

A simple way to realize such a function is to use
$
\ell(i,j,h) := \bar{l}(i,j,h) 
$, 
where~$\bar{l}(i,j,h)$ is defined as the average load reduction that has been previously obtained at daytime~$h$ by actor~$i$ after having applied action~$j$. In case no experience is available for action~$j$ of actor~$i$ an initial value is used, so
$
\ell(i,j,h) := I(i,j,h)
$
where~$I(i,j,h)$ is a user-defined optimistic initial value. To control the amount of exploration, we introduce a parameter called \emph{initial value weight}~$\beta \geq 0$, which determines how much the optimistic initial value is taken into account in later episodes.
In the computation of the average, each sample from real experience is weighted with $1$ while the initial value is weighted with $\beta$.


As another means of exploration we use the well-known~$\epsilon$-greedy scheme and apply it to each actor independently, that is, in each episode each actor ignores with probability~$\epsilon$ the assignment computed by the optimization module and applies instead an action selected uniformly at random.
A final parameter driving exploration for SE is the number~$\tau \geq 0$ of \emph{initial exploration episodes}, during which each actor applies random actions to obtain some initial experience.

\paragraph{Multi Episode algorithm}

A limitation of single episode algorithms is that in general it is impossible to adequately represent the population of all episodes by a single one. Solving the linear program using the expected values of~$\ell$ as parameters does in general not result in the action assignment maximizing the expected reward due to the nonlinear objective function in Equation~\ref{eq:objective}.  

One way to deal with this issue is to compute an assignment that is optimal for an ensemble of episodes. Given bias-free sample episodes~$1 \ldots N$, the average reward of assignment~$a$ is a bias-free and consistent estimator of the expected reward of~$a$. Thus, the assignment with maximal average reward among the $N$ samples almost surely converges against the optimal assignment.

Motivated by this observation, our \emph{Multi Episode (ME)} algorithm constructs a set of~$N$ sample episodes and applies the optimization module accordingly. 
Let~$S(i,j)$ be the set of samples collected for actor~$i$ and action~$j$, where each 
sample represents an~$H$-dimensional vector of load reduction values that have been observed from actor~$i$ having applied action~$j$. 
If no such sample has been collected yet, the set is defined using the (optimistic) initial values~$I(i,j,1) \ldots I(i,j,H)$.

A sample episode can be constructed by setting~$\ell_t(i,j,\cdot)$ to some value chosen from the sample set~$S(i,j)$. Note that for any~$i,j,i^\prime,j^\prime$ the values~$\ell_t(i,j,\cdot)$ and~$\ell_t(i^\prime,j^\prime,\cdot)$ possibly have been collected at different episodes. Thus, information on the correlation of the two curves is discarded by this construction method, and in general the sample episode is only bias-free under assumption of stochastic independence of the outcomes of all actor/action combinations.

We use the number of sample episodes~$N$ as a parameter of the Multi Episode algorithm, and we construct the~$N$ episodes by randomly picking from each~$S(i,j)$. After an element of~$S(i,j)$ has been selected, it is temporarily removed from the set until all other elements of~$S(i,j)$ have been selected (and removed) as well. This way we ensure that the available sample curves are distributed among the~$N$ sample episodes as evenly as possible, which reduces the correlation between the episodes and thus decreases the variance of the expected reward estimators.


\paragraph{Offline computation of optimal assignments}

To calculate the regret of the above algorithms we implement a method to compute offline an assignment very close to the optimum. A true sample episode can be obtained from simulated actors by querying their hypothetical behavior for each action. The assignment is computed by the optimization module using a sufficient number of such samples.

\section{Experiments}
\label{sec:experiments}

\paragraph{Experimental setup}

The~$n$ actors in our setup are modeled as~$n$ energy consumers, where the overall objective is to sustain the largest possible load reduction over a given daily \emph{target time interval}, which is discretized into daytime slots~$1,\ldots,H$; see Equation~\ref{eq:objective}. The actors interpret each action as the request to reduce energy usage during some \emph{request interval} which is a sub-interval of~$[1,H]$.
%
Out of the~$\Omega(H^2)$ possible sub-intervals we define an action set~$A$ of size~$O(H)$ as the smallest set with $[1,H] \in A$ and $[h_1,h_2] \in A \implies [h_1, \lceil\frac{1}{2}(h_1+h_2)\rceil], [\lceil\frac{1}{2}(h_1+h_2)\rceil,h2] \in A$.
%
%
Our consumer model mixes three types of loads.
Assuming that  for every consumer a baseline load curve has been fixed, we directly model load reductions against that hypothetical baseline. 

We model 
\emph{unconditional load reductions} as a discrete Gaussian process that is independent from applied actions. For each daytime~$h$ and actor~$i$, the unconditional load reduction~$u(i,h)$ is a Normal-distributed random variable with mean~$0$ and standard deviation~$\sigma_u$.
For some fixed~$z \in [0,1]$ and each daytime~$h>1$,~$u(i,h)$ is generated by~$u(i,h) = z \cdot u(i,h-1) + (1-z) \cdot \mathcal{N}(0,\sigma_u^2)$. From the linearity of mean and standard deviation of Gaussian variables follows that~$u(i,h) \sim \mathcal{N}(0,\sigma_u^2)$ for any~$h=1,\ldots,H$. In our experiments we work with fixed~$z=0.5$ but with varying values of~$\sigma_u$.

In addition, each consumer is assumed to have some \emph{curtailable load}, which represents energy consumption the consumer will suspend during the request interval. This curtailable load is generated uniformly at random as a constant value between 0 and 200W for each consumer.

Finally, every consumer has some \emph{shiftable load}, which is specified as a load with a specific daily starting time, length, and magnitude. The magnitude is generated as a uniform random number between 0.5 kW and 1.0 kW. The length~$L$ is a uniform random number between~$0.25 H$ and~$0.5 H$, and the start time is generated uniformly at random between time~$-0.5L$ and~$H-0.5L$, such that up to 50\% of the shiftable load can take place outside the target interval~$[1,H]$. This load can be freely shifted back and forth in time under the constraint that it remains within the union of its original time window and the target interval, and the consumer always tries to shift it such that its overlap with the request interval is minimized. Each consumer is with~50\% probability cooperative enough to curtail its shiftable load when its overlap with the request interval cannot be reduced.


Summarizing the model, each actor aggregates three distinct types of load reduction potentials with different characteristics. Note that the only prior information about this model available to the learning agent is an upper bound on the expected load reduction potential of~$2kW$ per actor.

\paragraph{Results}

All experiments were performed with a horizon of 365 episodes, corresponding to one year in our application. Linear optimization was done via the \texttt{pulp}~\cite{mitchell11} library for Python using \texttt{COIN-OR}~\cite{lougee03} as the backend optimizer. A runtime limit for the optimizer was set to 30s per episode, which turned out sufficient for a remaining optimality gap almost always under 5\%. For computing the optimal assignment offline, 20 sample days were taken from the simulation and the optimizer was given enough runtime for an optimality gap of less than 1\%. The algorithms were assigned an initial value of 2 kW as the load reduction potential for each actor, action, and daytime.

\begin{figure}
\begin{minipage}{0.45 \textwidth}
\includegraphics[width=\textwidth]{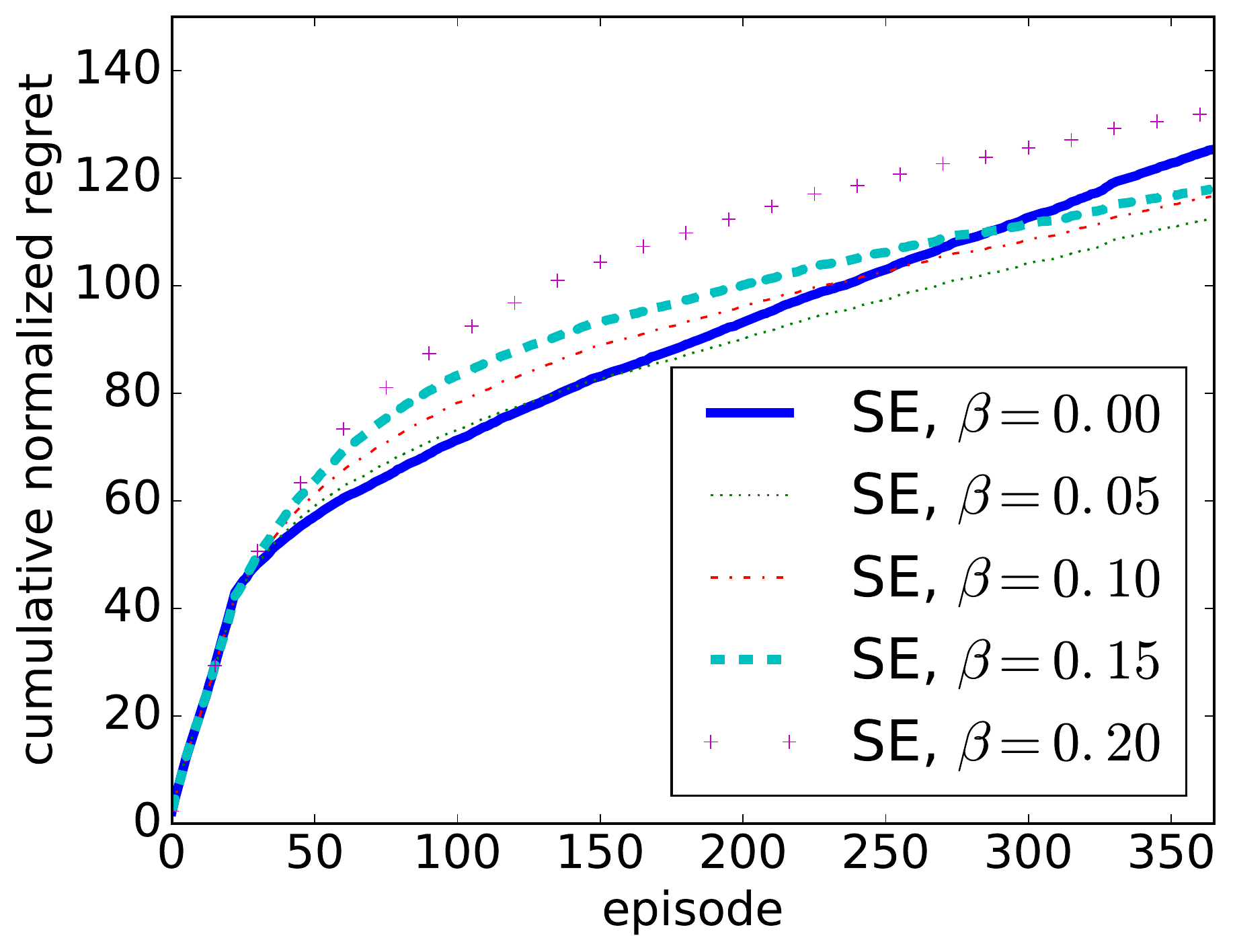}
\caption{
\label{fig:SE-beta}
SE algorithm with~$\beta \in [0,0.2]$,~$\epsilon=0$,~$\tau = 0$; simulation with~$\mu_u=500W$.
}
\end{minipage}
\hfill
\begin{minipage}{0.45 \textwidth}
\includegraphics[width=\textwidth]{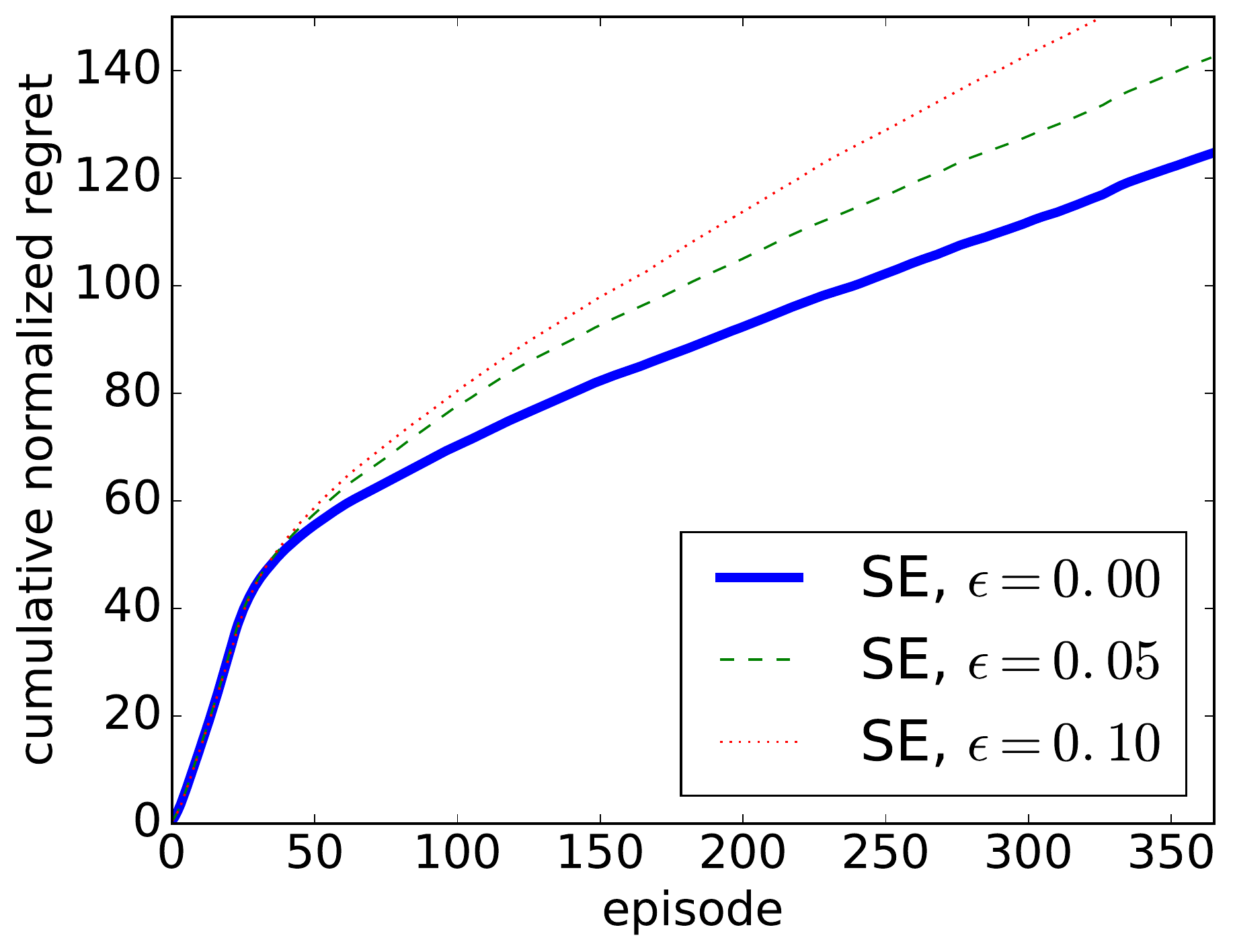}
\caption{
\label{fig:SE-epsilon}
SE algorithm with~$\beta = 0$,~$\epsilon\in[0,0.1]$,~$\tau = 0$; simulation with~$\mu_u=500W$.
}
\end{minipage}
\end{figure}

\begin{figure}
\begin{minipage}{0.45 \textwidth}
\includegraphics[width=\textwidth]{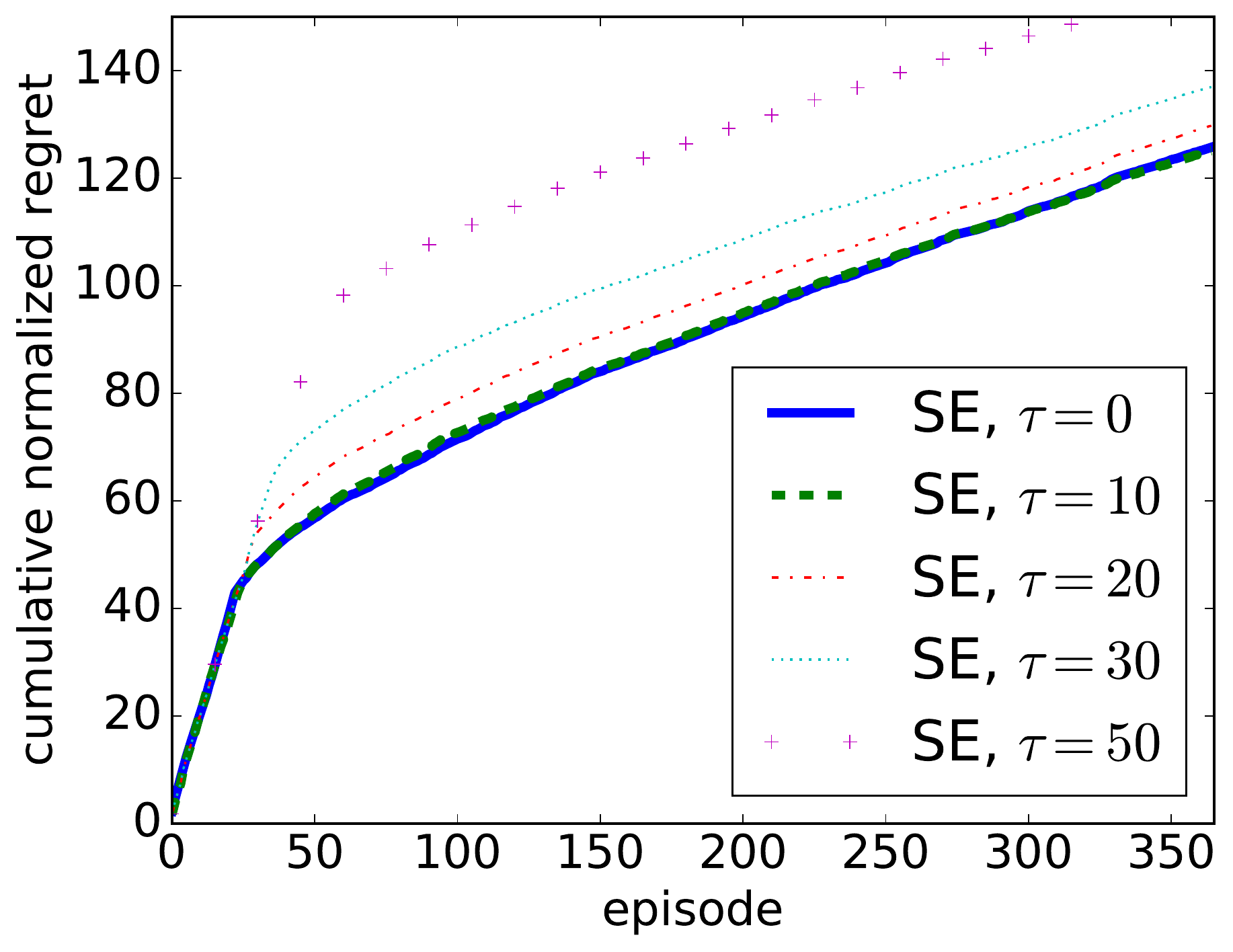}
\caption{
\label{fig:SE-tau}
SE algorithm with~$\beta = 0$,~$\epsilon=0$,~$\tau \in [0,50]$; simulation with~$\mu_u=500W$.
}
\end{minipage}
\hfill
\begin{minipage}{0.45 \textwidth}
\includegraphics[width=\textwidth]{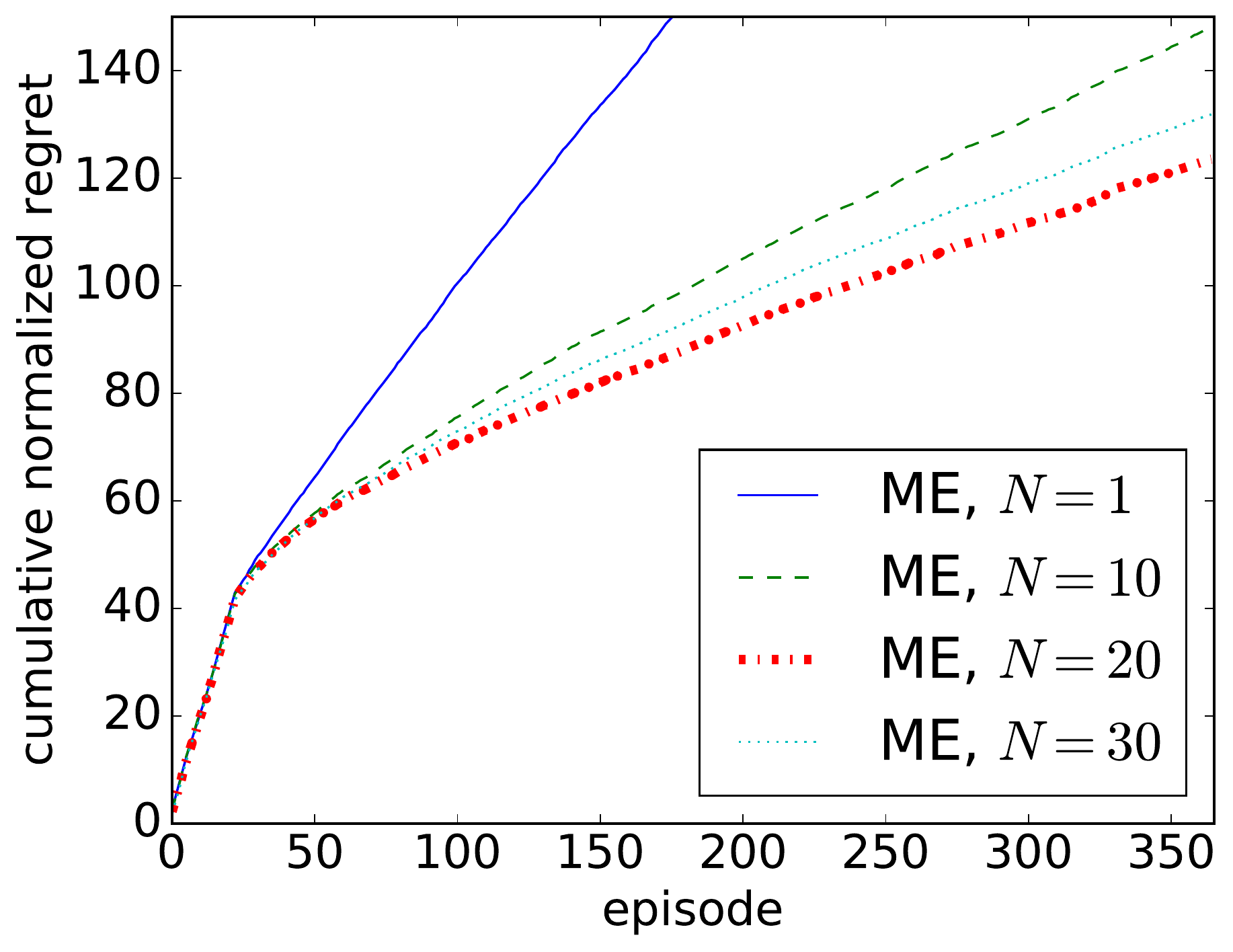}
\caption{
\label{fig:ME-N}
ME algorithm with~$N \in [1,30]$; simulation with~$\mu_u=500W$.
}
\end{minipage}
\end{figure}

In a first set of experiments the parameter spaces of the Single Episode (Figure~\ref{fig:SE-beta}-\ref{fig:SE-tau}) and Multi Episode (Figure~\ref{fig:ME-N}) algorithm are explored under the condition of rather large daily load curve fluctuations with~$\mu_u=500W$, making it difficult for learners to determine optimal assignments. 

Due to the optimistic initialization, all 24 actions are tried at least once by each actor, resulting in a steep regret curve during that phase regardless of the algorithm and its configuration. The influence of parameter~$\beta$, controlling for SE the amount of remaining optimism after an action has been taken once, exhibits the typical behavior of a parameter trading off exploration and exploitation, as observable in Figure~\ref{fig:SE-beta}. Setting $\beta=0$ is beneficial in early episodes, while values around $0.1$ pay off later.
As Figure~\ref{fig:SE-epsilon} shows, the well-known~$\epsilon$-greedy strategy does not improve exploration in the context of our experiments. Plausible reasons are that (a)~$\epsilon$-greedy does not distinguish between slightly suboptimal and clearly suboptimal actions, and (b) with 150 actors in each round the probability is high that some of them apply an action which does not combine well with the actions of the others. We set~$\epsilon=0$ in the remaining experiments. The final parameter of SE is~$\tau$, the number of initial random exploration episodes; see Figure~\ref{fig:SE-tau}. Also here it turns out that only for moderate values the initially larger regret has the potential to pay off within a reasonable number of episodes. 

For the Multi Episode algorithm we evaluate in Figure~\ref{fig:ME-N} the influence of the number of sample episodes~$N$ on the result. It turns out that the best performance is achieved for~$N=20$. A possible reason why the performance again deteriorates for~$N>20$ is the limited number of available samples per action. Building too many episodes from the same sample set at some point only increases the correlation among episodes and does not improve the reward estimator anymore. Furthermore, with more sample episodes it becomes computationally harder to determine a good action assignment within our runtime limit of 30s, thus the overall solution quality decreases.
\begin{figure}
\begin{minipage}{0.45 \textwidth}
\includegraphics[width=\textwidth]{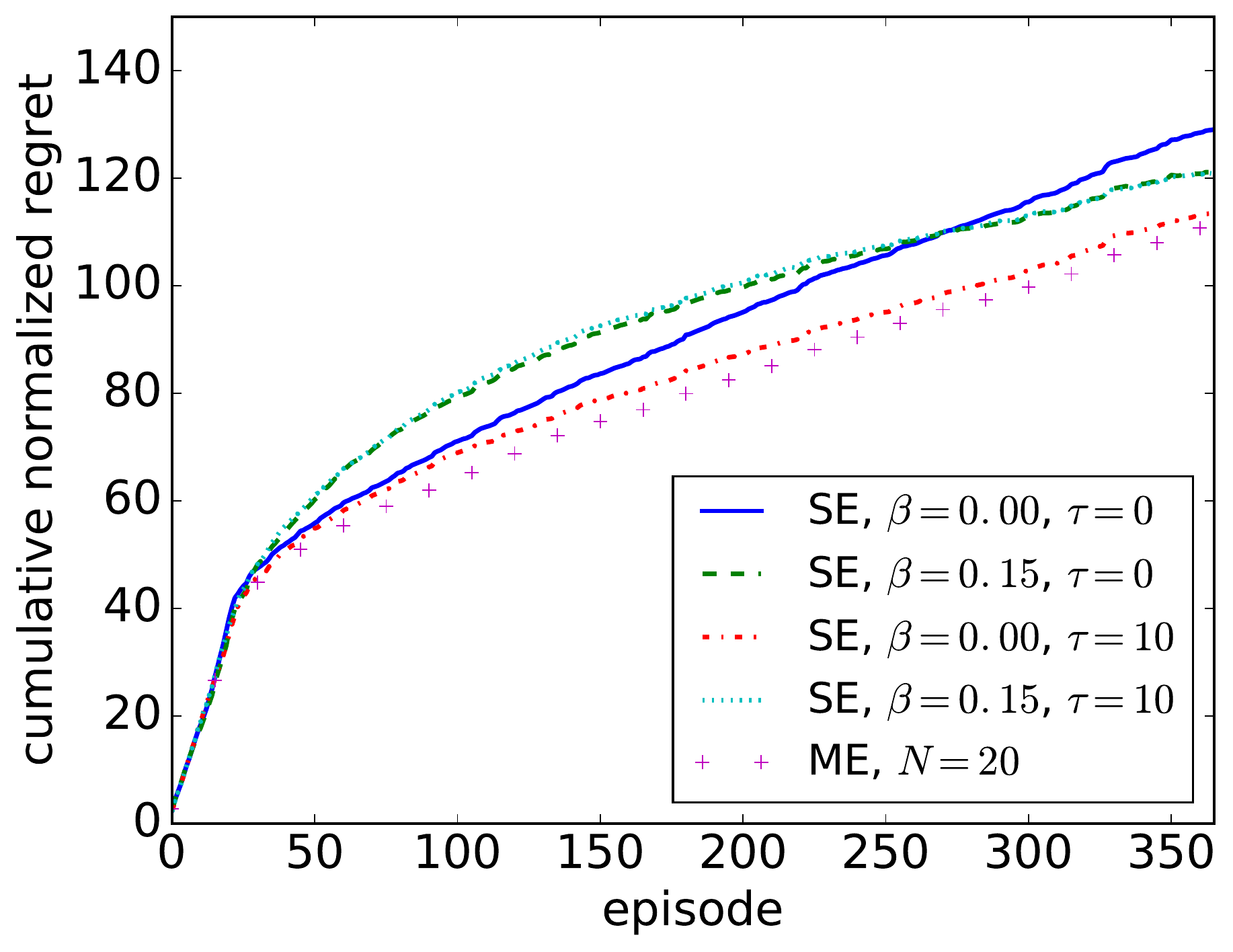}
\caption{
SE and ME algorithm applied to simulation with~$\mu_u=500W$.
\label{fig:tuned-500}
}
\end{minipage}
\hfill
\begin{minipage}{0.45 \textwidth}
\includegraphics[width=\textwidth]{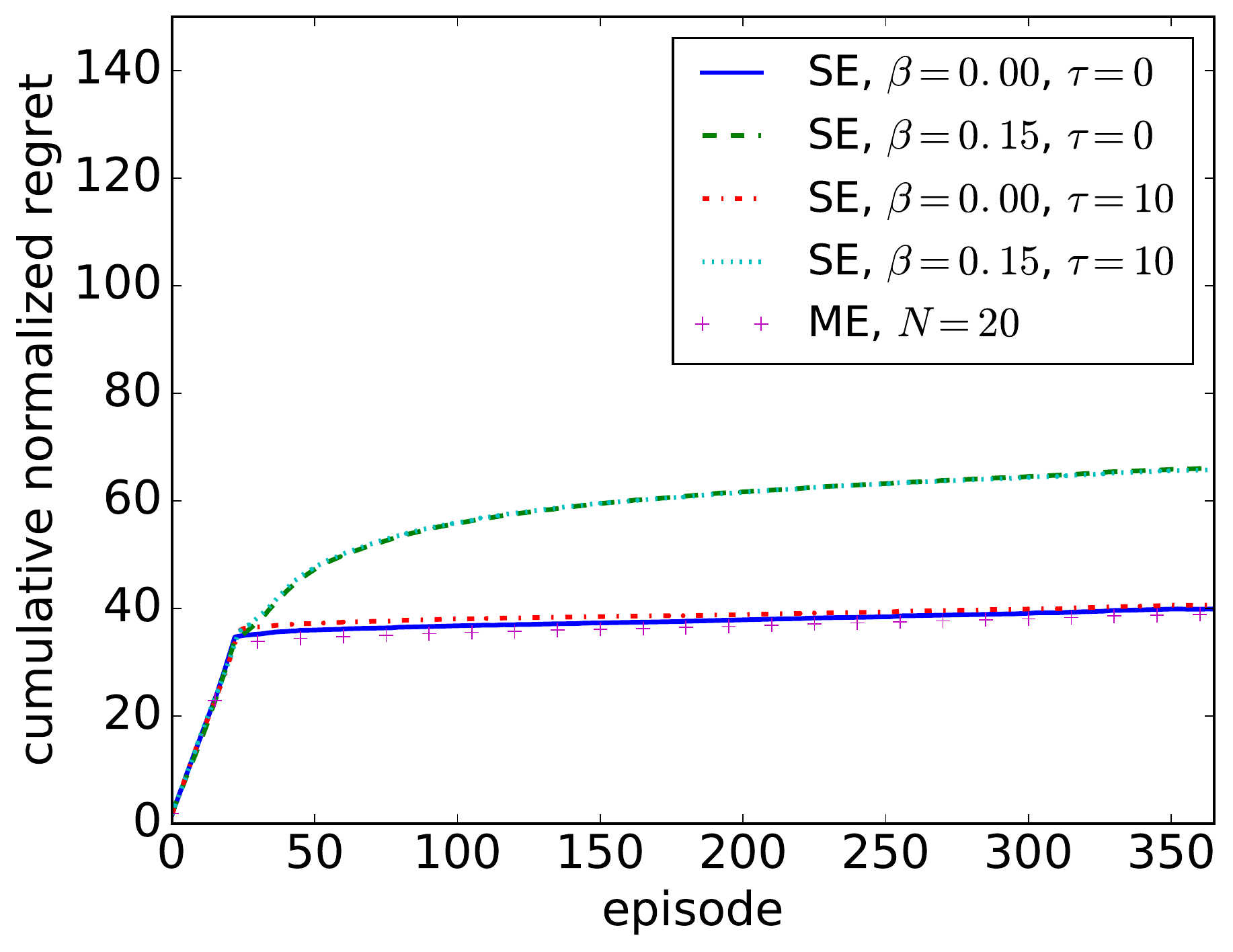}
\caption{
SE and ME algorithm applied to simulation with~$\mu_u=100W$.
\label{fig:tuned-100}
}
\end{minipage}
\end{figure}

In a final series of experiments we apply SE and ME using the most promising parameter configurations. The results for simulations with~$\mu_u = 500W$ and~$\mu_u = 100W$ are visualized in Figure~\ref{fig:tuned-500} and~\ref{fig:tuned-100}, respectively. The first observation is that there is no fundamental advantage of using multiple episodes in our application; combining SE with some initial exploration seems to lead to a similar performance. An interesting observation can be made for the Single Episode algorithm using optimism ($\beta=0.15$) to drive exploration. While slowing down convergence to the optimal assignment in case of low load fluctuations ($\mu_u = 100W$), it seems to be the only mechanism which continues to significantly improve the assignment over the course of all 365 episodes.  For~$\mu_u = 500W$, at the end of the curve in Figure~\ref{fig:tuned-500}, the curves with~$\beta=0.15$ have the least slopes among all tested algorithms, i.e. they apply the action assignments closest to the optimum. In general, Figure~\ref{fig:tuned-100} shows that both ME and SE can find action assignments very close to the optimum in environments with little randomness.

\section{Summary and open problems}

The Combinatorial Multi-Bandit Problem is a Reinforcement Learning setup with multiple actors. As our theoretical analysis has shown, the best trade-off between exploration and exploitation for instances of this problem is limited by regret~$\Theta(\log n)$, where the constant factor is determined by the regret of a combination of action assignments that sufficiently covers the suboptimal actions of all actors. The algorithms designed for our energy management application are based on a combination of bandit exploration schemes with mathematical programming. With minimal prior knowledge about the 24 actions of 150 actors in environments with high random fluctuations, the implementations achieve a cumulative normalized regret of around 115 in 365 episodes, which corresponds to about~$70\%$ of the optimal reward on average.

Further theoretical analysis will be required to derive upper regret bounds. The algorithms studied in our experiments have been tailored to a specific class of applications, and it is an open question whether these or other action selection rules can match the lower bound in general. In addition, generalizing the problem to stateful systems would substantially enhance its range of applications.

\bibliographystyle{plain}
\bibliography{RL4DR}

\end{document}